\documentclass[conference]{IEEEtran}
\IEEEoverridecommandlockouts
\usepackage{cite}
\usepackage{amsmath,amssymb,amsfonts}
\usepackage{algorithmic}
\usepackage{graphicx}
\usepackage{textcomp}
\usepackage{xcolor}
\usepackage{booktabs}
\usepackage[hidelinks]{hyperref}
\def\BibTeX{{\rm B\kern-.05em{\sc i\kern-.025em b}\kern-.08em
    T\kern-.1667em\lower.7ex\hbox{E}\kern-.125emX}}
\begin{document}

\title{DelowlightSplat: Feed-Forward Gaussian Splatting for Lowlight 3D Scene Reconstruction}

\author{\IEEEauthorblockN{1\textsuperscript{st} Fuzhen Jiang$^{*}$}
\IEEEauthorblockA{\textit{Hangzhou Dianzi University}\\
Hangzhou, China \\
$^{*}$23060111@hdu.edu.cn}
\and
\IEEEauthorblockN{2\textsuperscript{nd} Zengtian Xie}
\IEEEauthorblockA{\textit{Zhuhai College of Science and Technology} \\
Zhuhai, China \\
2598907113@stu.zcst.edu.cn}
\and
\IEEEauthorblockN{3\textsuperscript{rd} Zhuoran Li}
\IEEEauthorblockA{\textit{Hangzhou Dianzi University} \\
Hangzhou, China \\
23062207@hdu.edu.cn}
}

\maketitle

\begin{abstract}
Novel-view synthesis and 3D reconstruction from sparse posed images are central to robotics and AR/VR. Yet, feed-forward 3D Gaussian reconstruction fails under lowlight due to noise, color shifts, and unreliable correspondence. We propose DelowlightSplat, a lowlight-aware feed-forward Gaussian splatting framework for clean novel-view rendering. We build a controllable multi-view lowlight benchmark by degrading only context views while keeping target views clean. We introduce a lightweight Lowlight Adapter for residual enhancement to improve matchability, and couple it with cost-volume-based multi-view inference to directly predict clean 3D Gaussians. Experiments show that DelowlightSplat significantly outperforms previous feed-forward method and two-stage pipeline under lowlight conditions.
\end{abstract}


\section{Introduction}






Recovering a photorealistic 3D scene from a few posed images is fundamental to novel view synthesis. 3D Gaussian Splatting (3DGS) enables efficient, high-quality rendering \cite{b1}, while recent feed-forward methods aim to amortize per-scene optimization by predicting 3D Gaussians directly from sparse posed views \cite{b2}\cite{b10}\cite{b12}\cite{b13}.

Despite their efficiency, existing feed-forward pipelines are typically trained and evaluated on well-exposed imagery. Recent efforts on adverse-condition 3D reconstruction have shown increasing interest in degradation-aware Gaussian splatting, including low-resolution sparse-view reconstruction, smoke-degraded novel-view synthesis, and real-world benchmarks involving extreme low-light and smoke-corrupted inputs \cite{b14}\cite{b15}\cite{b16}. Prior work has also shown that novel-view synthesis becomes substantially harder in lowlight capture regimes \cite{b11}. In practice, reduced contrast, sensor noise, and color bias degrade feature matchability, which in turn breaks geometric reasoning based on cost-volume aggregation \cite{b3}, leading to incorrect geometry and over-smoothed appearance.

A straightforward workaround is to first enhance each lowlight view using a strong 2D restoration model and then run a feed-forward reconstructor. However, per-image enhancement may introduce view-dependent color/illumination shifts and does not explicitly optimize for multi-view consistency, so the downstream reconstruction can still be unstable.

We therefore target \emph{feed-forward 3DGS reconstruction under lowlight} and address two key obstacles. First, realistic lowlight multi-view supervision with clean targets is scarce. Second, directly applying a reconstructor trained on normal-exposure data generalizes poorly to lowlight inputs. To bridge the data gap, we construct a controllable benchmark by degrading only the context views (gamma darkening, exposure scaling, RGB channel shift, and blur) while keeping target views clean, based on a large-scale posed-view dataset \cite{b4}.

Built upon this benchmark, we propose \textbf{DelowlightSplat}, a two-view feed-forward 3DGS reconstructor for lowlight inputs. It integrates a lightweight \emph{Lowlight Adapter} to perform residual enhancement tailored for multi-view matching, and a cost-volume-based multi-view encoder to infer geometry and appearance and directly predict a clean, normally illuminated 3D Gaussian scene for novel-view rendering.

Our contributions are summarized as follows:
\begin{itemize}
  \item \textbf{A Controllable Multi-view Lowlight Benchmark.} We build a training/evaluation benchmark by degrading only context views with probabilistic lowlight transformations while keeping target views clean.
  \item \textbf{A Novel Delowlight Pipeline.} We propose a two-view lowlight-aware feed-forward 3DGS pipeline \textbf{DelowlightSplat} that improves cross-view matchability via a lightweight Lowlight Adapter and predicts clean 3D Gaussians for direct novel-view rendering.
\end{itemize}

\begin{figure*}[!t]
    \centering
    \includegraphics[width=0.8\textwidth]{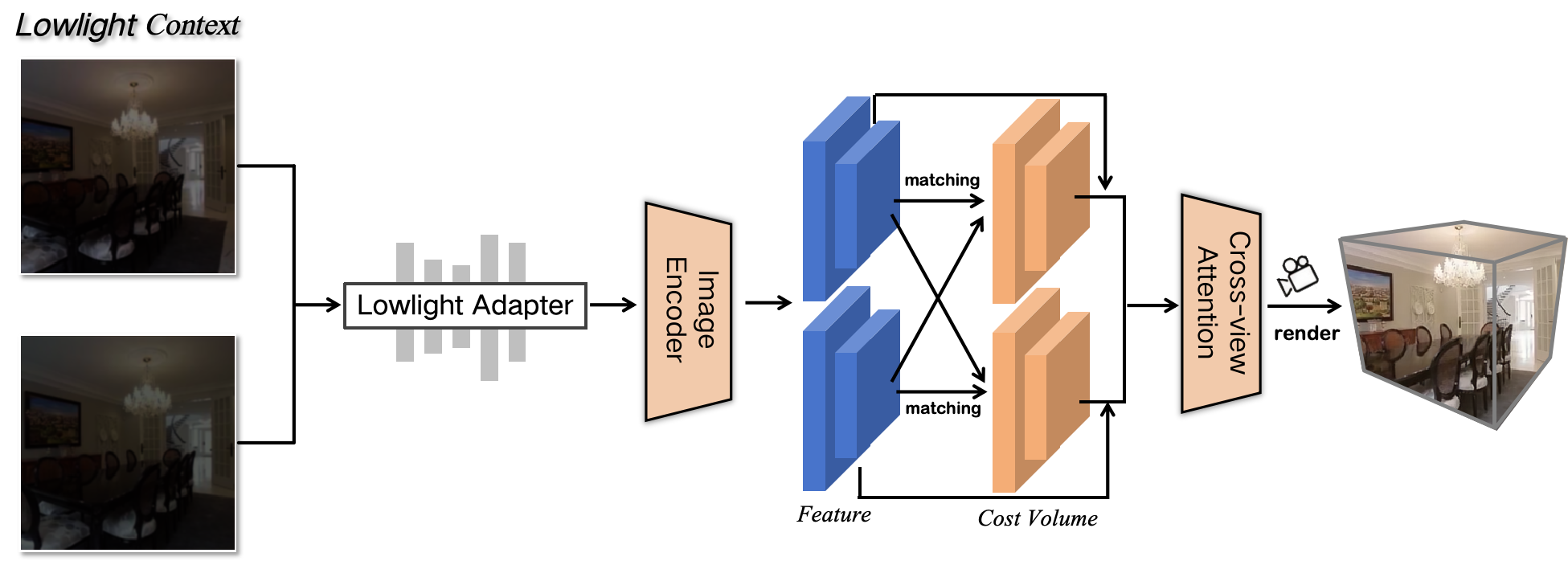}
    \caption{\textit{Overall pipeline of DelowlightSplat}. Lowlight context views are first adapted by a lightweight lowlight adapter, then fused by cost-volume-based multi-view inference to predict a clean 3D Gaussian scene for novel-view rendering.}
    \label{fig:pipeline}
\end{figure*}

\section{Related Work}

\subsection{Feed-forward 3D Gaussian Reconstruction}
3DGS provides an explicit, efficient scene representation for real-time novel-view rendering \cite{b1}, but standard pipelines still optimize Gaussians per scene. Feed-forward reconstruction amortizes this process by predicting Gaussian primitives from sparse posed views \cite{b2}, often using cost-volume aggregation to encode multi-view consistency \cite{b3}. Our work follows this paradigm and focuses on improving feature matchability and geometric reliability in low-light conditions.

\subsection{2D Lowlight Image Enhancement}
Lowlight Image Enhancement (LLIE) methods map lowlight inputs to normal-exposure images using paired/synthetic supervision (e.g., Retinex-style decomposition \cite{b5} and raw-to-RGB training \cite{b6}), as well as weak/unsupervised objectives such as zero-reference curve estimation \cite{b7} and unpaired adversarial learning \cite{b8}. Recent efficient restoration models, such as DarkIR \cite{b9}, further improve robustness by jointly addressing under-exposure, noise, and blur.

However, most LLIE is applied per image and may break cross-view consistency (e.g., color drift or hallucinations), which can harm multi-view reconstruction. In our experiments, we also consider a strong ``restore-then-reconstruct'' baseline: enhancing each lowlight view using DarkIR \cite{b9} and then feeding the restored images into a feed-forward 3DGS reconstructor MVSplat \cite{b10}. By contrast, our method integrates lowlight adaptation into the reconstruction pipeline to explicitly improve multi-view matchability and produce a clean 3D Gaussian scene.

\section{Methodology}

\subsection{Overview}
DelowlightSplat targets feed-forward 3D reconstruction and novel-view synthesis from \emph{low-light} two-view inputs. Fig.~\ref{fig:pipeline} overviews the pipeline. Given two low-light context views with known camera calibration, our goal is to directly predict a renderable 3DGS scene representation and synthesize \emph{clean, normally illuminated} images from arbitrary target viewpoints.

We predict a set of Gaussians
\begin{equation}
\mathcal{G}=\{(\boldsymbol{\mu}_j,\boldsymbol{\Sigma}_j,\mathbf{h}_j,\alpha_j)\}_{j=1}^{N_g},
\end{equation}
where $\boldsymbol{\mu}_j\in\mathbb{R}^3$ denotes the 3D mean, $\boldsymbol{\Sigma}_j\in\mathbb{R}^{3\times 3}$ the covariance, $\mathbf{h}_j$ the spherical harmonics (SH) coefficients for view-dependent color, and $\alpha_j\in(0,1)$ the opacity.
Given a target camera $(\mathbf{K}_t,\mathbf{T}_t)$, a differentiable Gaussian splatting renderer produces the 
\begin{equation}
\hat I_t=\mathcal{R}(\mathcal{G};\mathbf{K}_t,\mathbf{T}_t).
\end{equation}

The overall workflow is:
\emph{Lowlight Context} $\rightarrow$ \emph{Lowlight Adapter} $\rightarrow$ \emph{Multi-view Geometric/Appearance Inference} $\rightarrow$ \emph{3DGS Construction} $\rightarrow$ \emph{Splatting-based Rendering}.

\subsection{Lowlight Dataset}
\label{sec:lowlight_dataset}
Obtaining real lowlight multi-view captures with accurate calibration and paired clean ground truth is challenging. We therefore degrade only the context views while keeping the target views clean.
Formally,
\begin{equation}
I_i^{\text{low}}=\mathcal{L}(I_i^{\text{clean}}),\qquad
I_t^\star=I_t^{\text{clean}}.
\end{equation}

This protocol isolates how lowlight inputs affect reconstruction while preventing the model from simply imitating lowlight appearance. Instead, the objective is to recover a clean and consistent scene from lowlight observations.

The lowlight operator $\mathcal{L}$ composes four controllable transformations in sequence.
Each context image is independently degraded with probability $p$ (with $p{=}1$ meaning all context images are degraded).

\noindent \textbf{(i) Gamma Darkening.}
\begin{equation}
I \leftarrow I^\gamma,\qquad
\gamma \sim \mathcal{U}(\gamma_{\min},\gamma_{\max}),\ \gamma>1.
\end{equation}

\noindent \textbf{(ii) Exposure Scaling.}
\begin{equation}
I \leftarrow s\,I,\qquad
s \sim \mathcal{U}(s_{\min},s_{\max}),\ 0<s<1.
\end{equation}

\noindent \textbf{(iii) RGB Channel Shift.}
\begin{equation}
I_c \leftarrow (1+\delta_c)\,I_c,\qquad
\delta_c \sim \mathcal{U}(-r,r),\ c\in\{R,G,B\}.
\end{equation}

\noindent \textbf{(iv) Blur.}
\begin{equation}
I \leftarrow G_{\sigma,k} * I,
\end{equation}
where $G_{\sigma,k}$ is a Gaussian kernel with standard deviation $\sigma$ and kernel size $k$.

By adjusting $\gamma, s, r, \sigma, k,$ and $p$, the degradation strength is interpretable and reproducible, enabling systematic ablations to show how each lowlight effect influences reconstruction.

\begin{figure*}[!t]
    \centering
    \includegraphics[width=0.8\textwidth]{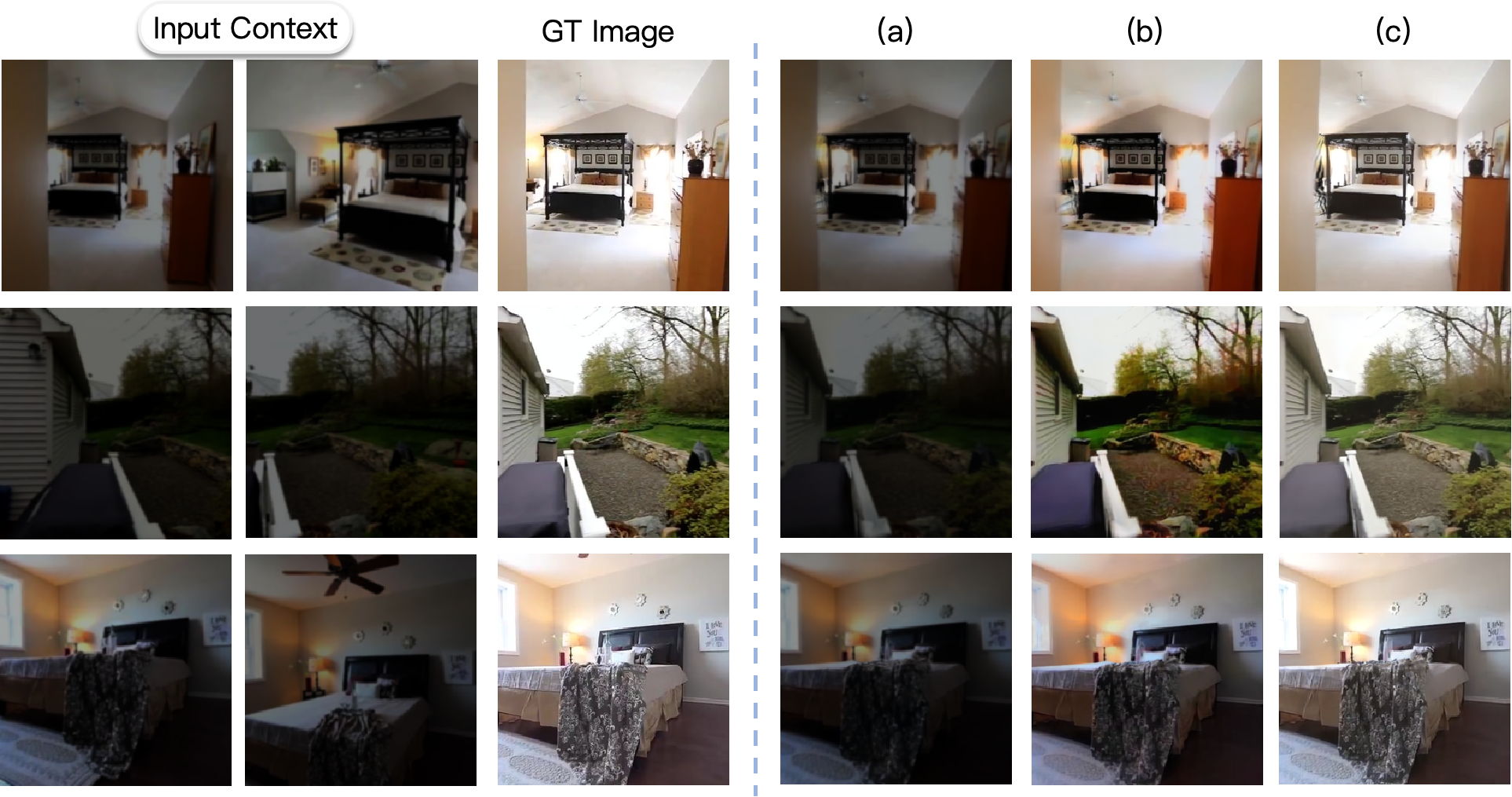}
    \caption{\textit{Qualitative comparison on lowlight novel-view synthesis.} (a) Direct reconstruction baseline (Lowlight MVSplat). (b) Restore-then-reconstruct baseline (DarkIR$\rightarrow$MVSplat). (c) \textbf{DelowlightSplat}. Our method yields sharper details and more consistent appearance across novel views.}
    \label{fig:qualitative}
\end{figure*}

\subsection{Lowlight Adapter}
Lowlight images exhibit insufficient brightness, color bias, and missing fine details, which directly weakens multi-view matching cues and destabilizes downstream geometry inference.
We introduce a lightweight front-end \emph{Lowlight Adapter} to map lowlight inputs into a representation domain that is more suitable for multi-view reasoning.

Given an input context image $I_i$, the adapter predicts a residual correction and produces an adapted image $\tilde I_i$:
\begin{equation}
\tilde I_i=\mathrm{clip}\big(I_i+\lambda\cdot\tanh(\Delta(I_i))\big),
\end{equation}
where $\Delta(\cdot)$ is implemented by a small stack of convolutional residual blocks, $\lambda$ controls the residual magnitude, and $\mathrm{clip}(\cdot)$ enforces valid intensity range.
This residual formulation has two practical benefits: (i) when degradation is mild, the mapping can stay close to identity to avoid over-enhancement; (ii) the adapter is computationally light and can be trained end-to-end jointly with the reconstruction backbone.

Importantly, the adapter is not designed as a standalone 2D ``brightening'' module. Instead, it is optimized to improve the matchability and geometric usefulness of representations for the subsequent multi-view pipeline.

\subsection{Feed-forward Gaussian Splatting Network}
Given the adapted context views $\{\tilde I_i\}$ and camera calibration, we predict a 3D Gaussian set $\mathcal{G}$ in a single forward pass and render it from target viewpoints.

\noindent \textbf{Multi-view feature encoding.}
We first extract 2D features from each context view, and then perform cross-view interaction to aggregate geometrically consistent evidence, yielding multi-view features that support both depth reasoning and appearance prediction.

\noindent \textbf{Depth-driven geometry construction.}
At each pixel location, the network predicts depth along with density-related quantities, and back-projects image measurements into 3D to form Gaussian centers:
\begin{equation}
\boldsymbol{\mu}(\mathbf{u},d)=\mathbf{o}+d\,\mathbf{r}(\mathbf{u};\mathbf{K},\mathbf{T}),
\end{equation}
where $\mathbf{u}$ is the pixel coordinate, $\mathbf{o}$ is the camera center, and $\mathbf{r}(\mathbf{u};\mathbf{K},\mathbf{T})$ is the corresponding world-space ray direction.
Covariances $\boldsymbol{\Sigma}$ are parameterized from predicted scale and rotation to ensure positive definiteness, and opacities are bounded to $(0,1)$ via a squashing nonlinearity.

\noindent \textbf{Renderable appearance via spherical harmonics.}
Each Gaussian carries SH coefficients $\mathbf{h}$ to represent view-dependent color, allowing a single 3DGS to render consistent appearance across target viewpoints.

\noindent \textbf{Training objective.}
For each target camera $(\mathbf{K}_t,\mathbf{T}_t)$, we render $\hat I_t$ and supervise it with the clean target $I_t^\star$ using a combination of pixel-wise and perceptual losses:
\begin{equation}
\mathcal{L}=\mathcal{L}_{\text{pix}}(\hat I_t,I_t^\star)+\lambda_{\text{per}}\mathcal{L}_{\text{per}}(\hat I_t,I_t^\star),
\end{equation}
where $\mathcal{L}_{\text{pix}}$ can be $\ell_2$ or Charbonnier loss, and $\mathcal{L}_{\text{per}}$ denotes a perceptual loss.
At inference time, the model takes only low-light context views and the target camera pose(s), and outputs clean novel views through a single forward pass and Gaussian splatting rendering.

\section{Experiments}

\subsection{Settings}

\noindent \textbf{Baselines.}
We compare DelowlightSplat with three baselines:

\textit{(i) Lowlight MVSplat.} We directly feed the lowlight context views into MVSplat \cite{b10} without any enhancement, to quantify how lowlight degradations affect feed-forward gaussian reconstruction in our lowlight setting.

\textit{(ii) DarkIR $\rightarrow$ MVSplat.} We first restore each lowlight view using DarkIR \cite{b9} and then run MVSplat \cite{b10} on the restored images, representing a strong two-stage ``restore-then-reconstruct'' pipeline.

\textit{(iii) \textbf{DelowlightSplat~(Ours)}.} Our method performs lightweight lowlight adaptation before multi-view inference and predicts a clean, normally illuminated 3D Gaussian scene end-to-end for novel-view rendering.

\noindent \textbf{Implementation Details.}
During training, we adopt a curriculum on the context-view degradation by gradually increasing the degradation probability $p$, which stabilizes early optimization and improves robustness under varying low-light severity. All experiments are run on an NVIDIA RTX 4090 GPU with batch size 8 for 10,000 training steps.

\subsection{Main Results}

\noindent \textbf{Qualitative Results.}
As shown in Fig.~\ref{fig:qualitative}, feeding lowlight views directly into MVSplat produces over-smoothed textures and occasional geometric artifacts (e.g., floating structures) due to unreliable correspondences. DarkIR$\rightarrow$MVSplat baseline improves overall brightness, but the per-image restoration can introduce view-dependent color/illumination shifts that reduce multi-view consistency and lead to residual blur in novel views. In contrast, \textbf{DelowlightSplat} yields sharper edges and more stable geometry, with more consistent color and shading across viewpoints.

\begin{table}[t]
\caption{Quantitative comparison on low-light novel-view synthesis.}
\label{tab:main_baselines}
\centering
\begin{tabular}{lccc}
\toprule
Method & PSNR $\uparrow$ & SSIM $\uparrow$ & LPIPS $\downarrow$ \\
\midrule
Lowlight MVSplat & 8.49 & 0.372 & 0.355 \\
DarkIR $\rightarrow$ MVSplat & \underline{17.11} & \underline{0.723} & \underline{0.225} \\
\textbf{Ours} & \textbf{23.08} & \textbf{0.833} & \textbf{0.162} \\
\bottomrule
\end{tabular}
\end{table}

\noindent \textbf{Quantitative Results.}
Tab.~\ref{tab:main_baselines} summarizes the results on low-light novel-view synthesis. Our approach improves PSNR from 8.49 to 23.08 (+14.59) and SSIM from 0.372 to 0.833 (+0.461) over Lowlight MVSplat, and further surpasses DarkIR$\rightarrow$MVSplat by +5.97 PSNR and +0.110 SSIM while reducing LPIPS from 0.225 to 0.162 (\,$\downarrow$0.063). These consistent gains indicate higher reconstruction fidelity and better perceptual quality under low light conditions.

\begin{figure}[t]
    \centering
    \includegraphics[width=0.9\linewidth]{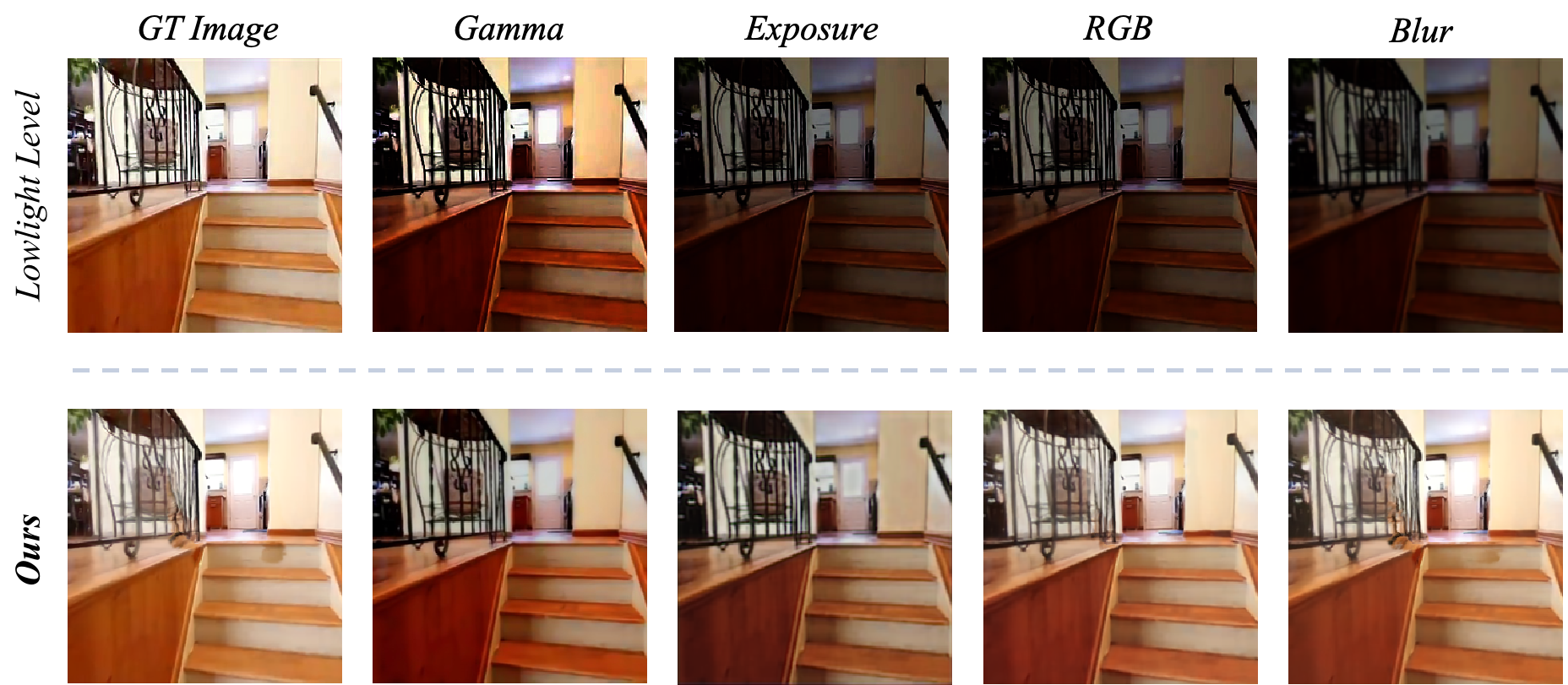}
    \caption{\textit{Step-wise visualization of the lowlight degradation pipeline} used to generate context views: Clean $\rightarrow$ Gamma Darkening $\rightarrow$ Exposure Scaling $\rightarrow$ RGB Channel Shift $\rightarrow$ Blur. The bottome row visualizing the corresponding results after applying our approach.}
    \label{fig:ablation}
\end{figure}

\begin{table}[t]
\caption{Quantitative ablation corresponding to Fig.~\ref{fig:ablation}. The reported scores are obtained by feeding each lowlight level into ours.}
\label{tab:ablation_fig3}
\centering
\begin{tabular}{lccc}
\toprule
Lowlight Level & PSNR $\uparrow$ & SSIM $\uparrow$ & LPIPS $\downarrow$ \\
\midrule
Clean & \textbf{25.07} & \textbf{0.852} & \textbf{0.144} \\
+ Gamma Darkening & 18.39 & 0.703 & 0.194 \\
+ Exposure Scaling & 20.38 & 0.787 & 0.286 \\
+ RGB Channel Shift & 21.29 & 0.788 & 0.285 \\
+ Blur & \underline{23.08} & \underline{0.833} & \underline{0.162} \\
\bottomrule
\end{tabular}
\end{table}

\subsection{Ablation Study}
We study how individual lowlight factors affect feed-forward 3DGS reconstruction by applying each degradation in Sec.~\ref{sec:lowlight_dataset} to the context views and evaluating our method (Fig.~\ref{fig:ablation} and Tab.~\ref{tab:ablation_fig3}). Overall, reduced illumination (gamma/exposure), color bias (RGB shift), and loss of high-frequency details (blur) weaken cross-view correspondence and thus destabilize cost-volume aggregation. Our lowlight adapter mitigates these issues by improving feature matchability, leading to more faithful and consistent reconstructions across all degradation types.

\section{Conclusion}
We introduced DelowlightSplat, a lowlight-aware feed-forward Gaussian splatting framework for novel-view synthesis and 3D reconstruction from sparse posed imagery. By coupling a lightweight residual Lowlight Adapter with cost-volume multi-view inference, the network restores matchable cues, stabilizes geometric aggregation, and directly predicts a clean, normally illuminated 3D Gaussian scene for rendering. Comparisons against direct reconstruction and restore-then-reconstruct baselines verify clear, consistent improvements, and perceptual sharpness under diverse lowlight degradations.

A promising direction for future work is to extend the model to more views and longer temporal contexts, and to improve robustness under real sensor artifacts such as severe motion blur and spatially varying noise.






\begin{thebibliography}{00}

\bibitem{b1} B. Kerbl, G. Kopanas, T. Leimk\"uhler, and G. Drettakis, ``3D Gaussian splatting for real-time radiance field rendering,'' \emph{ACM Trans. Graph.}, vol. 42, no. 4, Art. no. 139, pp. 1--14, 2023, doi: 10.1145/3592433.

\bibitem{b2} D. Charatan, S. L. Li, A. Tagliasacchi, and V. Sitzmann, ``pixelSplat: 3D Gaussian splats from image pairs for scalable generalizable 3D reconstruction,'' in \emph{Proc. IEEE/CVF Conf. Comput. Vis. Pattern Recognit. (CVPR)}, 2024, pp. 19457--19467.

\bibitem{b10} T. Huang, J. Wang, and Y. Wang, ``MVSplat: Efficient multi-view 3D Gaussian splatting,'' arXiv preprint arXiv:2403.14627, 2024.

\bibitem{b11} J. T. Barron, B. Mildenhall, M. Tancik, P. Hedman, R. Martin-Brualla, and P. Srinivasan, ``NeRF in the Dark: High dynamic range view synthesis from noisy raw images,'' in \emph{Proc. IEEE/CVF Conf. Comput. Vis. Pattern Recognit. (CVPR)}, 2022, pp. 16190--16199.

\bibitem{b12} Y. Wang, Z. Yan, P. Guo, Z. Wang, and Y. Gao, ``FreeSplat: Generalizable 3D Gaussian splatting for free-view synthesis of indoor scenes,'' arXiv preprint arXiv:2405.17958, 2024.

\bibitem{b13} S. Hong, H. Lee, W. Han, and H. Kim, ``PF3plat: Pose-free feed-forward 3D Gaussian splatting for novel view synthesis,'' in \emph{Proc. Int. Conf. Mach. Learn. (ICML)}, 2025, pp. 23662--23681.

\bibitem{b14} S. Liu, C. Bao, Z. Cui, X. Chu, B. Ren, L. Gu, X. Chen, M. Li, L. Ma, M. V. Conde, \emph{et al.}, ``NTIRE 2026 3D restoration and reconstruction in real-world adverse conditions: RealX3D challenge results,'' arXiv preprint arXiv:2604.04135, 2026, doi: 10.48550/arXiv.2604.04135.

\bibitem{b15} X. Hu, C. Shi, C. Yang, M. Chen, J. Ding, T. Wei, C. Wei, Z. Yu, and M. Tan, ``SRSplat: Feed-forward super-resolution Gaussian splatting from sparse multi-view images,'' in \emph{Proc. AAAI Conf. Artif. Intell.}, vol. 40, no. 6, pp. 4950--4958, 2026, doi: 10.1609/aaai.v40i6.42499.

\bibitem{b16} Q. Cao, X. Hu, C. Shi, J. Ding, Z. Yu, and J. Yu, ``GenSmoke-GS: A multi-stage method for novel view synthesis from smoke-degraded images using a generative model,'' arXiv preprint arXiv:2604.03039, 2026, doi: 10.48550/arXiv.2604.03039.

\bibitem{b3} Y. Yao, Z. Luo, S. Li, T. Fang, and L. Quan, ``MVSNet: Depth inference for unstructured multi-view stereo,'' in \emph{Computer Vision -- ECCV 2018}, V. Ferrari, M. Hebert, C. Sminchisescu, and Y. Weiss, Eds. Cham, Switzerland: Springer, 2018, pp. 785--801.

\bibitem{b4} T. Zhou, R. Tucker, J. Flynn, G. Fyffe, and N. Snavely, ``Stereo magnification: Learning view synthesis using multiplane images,'' \emph{ACM Trans. Graph.}, vol. 37, no. 4, Art. no. 65, 2018, doi: 10.1145/3197517.3201323.

\bibitem{b5} C. Wei, W. Wang, W. Yang, and J. Liu, ``Deep retinex decomposition for low-light enhancement,'' in \emph{Proc. Brit. Mach. Vis. Conf. (BMVC)}, 2018.

\bibitem{b6} C. Chen, Q. Chen, J. Xu, and V. Koltun, ``Learning to see in the dark,'' in \emph{Proc. IEEE/CVF Conf. Comput. Vis. Pattern Recognit. (CVPR)}, 2018, pp. 3291--3300.

\bibitem{b7} C. Guo, C. Li, J. Guo, C. C. Loy, J. Hou, S. Kwong, and R. Cong, ``Zero-reference deep curve estimation for low-light image enhancement,'' in \emph{Proc. IEEE/CVF Conf. Comput. Vis. Pattern Recognit. (CVPR)}, 2020, pp. 1780--1789.

\bibitem{b8} Y. Jiang, X. Gong, D. Liu, C. Yu, F. Chen, X. Shen, J. Yang, P. Zhou, and Z. Wang, ``EnlightenGAN: Deep light enhancement without paired supervision,'' \emph{IEEE Trans. Image Process.}, vol. 30, pp. 2340--2349, 2021, doi: 10.1109/TIP.2021.3051462.

\bibitem{b9} D. Feijoo, J. C. Benito, A. Garcia, and M. V. Conde, ``DarkIR: Robust low-light image restoration,'' in \emph{Proc. IEEE/CVF Conf. Comput. Vis. Pattern Recognit. (CVPR)}, 2025, pp. 10879--10889.

\end{thebibliography}
\end{document}